# Real-time Human Detection in Fire Scenarios using Infrared and Thermal Imaging Fusion

Nghe-Nhan Truong
*Faculty of Electrical and Electronics Engineering*
*HCMC University of Technology and Education*
Ho Chi Minh City, Vietnam
18161125@student.hcmute.edu.vn

Truong-Dong Do
*Dept. of Aerospace Systems Engineering*
*Sejong University*
Seoul, South Korea
dongdo@sju.ac.kr

My-Ha Le*
*Faculty of Electrical and Electronics Engineering*
*HCMC University of Technology and Education*
Ho Chi Minh City, Vietnam
halm@hcmute.edu.vn

*Abstract*— Fire is considered one of the most serious threats to human lives which results in a high probability of fatalities. Those severe consequences stem from the heavy smoke emitted from a fire that mostly restricts the visibility of escaping victims and rescuing squad. In such hazardous circumstances, the use of a vision-based human detection system is able to improve the ability to save more lives. To this end, a thermal and infrared imaging fusion strategy based on multiple cameras for human detection in low-visibility scenarios caused by smoke is proposed in this paper. By processing with multiple cameras, vital information can be gathered to generate more useful features for human detection. Firstly, the cameras are calibrated using a Light Heating Chessboard. Afterward, the features extracted from the input images are merged prior to being passed through a lightweight deep neural network to perform the human detection task. The experiments conducted on an NVIDIA Jetson Nano computer demonstrated that the proposed method can process with reasonable speed and can achieve favorable performance with a mAP@0.5 of 95%.

## I. INTRODUCTION

Fire accident is one of the most tragic occurrences that people have to deal with. The importance of detecting and rescuing victims in those situations cannot be overstated, as it can be a matter of life and death. According to the National Fire Protection Association, in 2019 alone, there were an estimated 1,291,500 fires in the United States, causing 3,704 civilian deaths, 16,600 civilian injuries, and $14.8 billion in property damage [1]. In addition to the direct impact on human lives and property, fire and smoke also have negative impacts on the environment and the economy [2], [3]. Therefore, there always exists the need to develop effective detection and rescue systems to minimize those unforeseen damages.

However, detecting and rescuing victims in fire accidents is a challenging mission due to the restricted visibility caused by smoke and debris. Traditional methods such as visual inspection and manual search are not only time-consuming, labor-intensive, and often ineffective, but also dangerous, directly threatening the safety of firefighters [4], [5]. This is where camera-based detection systems can be extremely useful. In recent years, cameras and sensors have become increasingly popular for detecting objects in fire search and rescue scenarios [6]. Thermal cameras and infrared (IR) cameras are commonly used for this purpose [7]. Nevertheless, each type of camera has its advantages as well as drawbacks. In an environment filled with substantial smoke, IR cameras lose clarity, reduce reliability, and even become blinded (Fig. 1a), whereas thermal cameras demonstrate their effectiveness since they gather the heat emitted from human bodies without being affected by smoke (Fig. 1c). However, if it had slight smoke, but there was a source of heat near enough to the people, as shown in (Fig. 1b), it would result in confusion for the thermal camera when trying to search for a human inside of a fire. IR camera information (Fig. 1d) can be used at this time to make decisions. Therefore, in such cases, a fusion approach combining the strengths of the two cameras can yield better detection results.

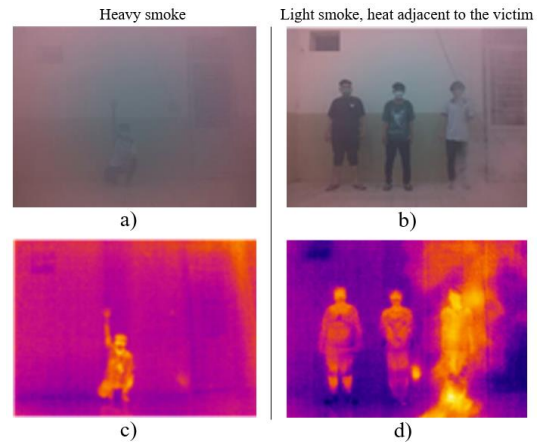

Fig. 1. Comparison between infrared and thermal cameras in two typical fire scenes

Multi-sensor fusion has been an important research area in recent years, especially in the field of computer vision. This technique combines the signals measured by multiple sensors to create a more complete and accurate description of the environment. There are several methods of multi-sensor fusion including data-level fusion, feature-level fusion, decision-level fusion, and sensor-level fusion. Among them, data-level fusion is the most used method, which directly fuses the raw data from multiple sensors [8], [9]. One common multi-sensor data fusion method is the fusion of thermal and RGB images, which have been used for human detection [10]. Another approach involves fusing data from multiple sensors at the decision-level, such as combining texture and color features extracted from multimodal data [11]. However, conventional multi-sensor fusion methods often rely on hand-crafted features, which can be time-consuming and may not generalize well to different environments. Recently, deep learning-based methods have emerged as a promising approach for multi-sensor fusion. For example, a deep neural network has been proposed for fusing data from multiple

---

* *Corresponding author.*

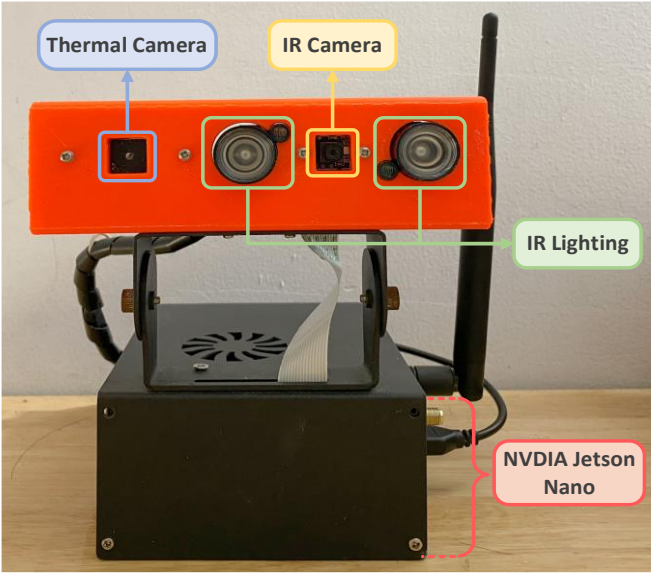

Fig. 2. The hardware for human detection consists of infrared and thermal cameras

sensors, including camera data and projected sparse radar, for object detection [12].

Deep neural networks (DNNs) have achieved remarkable success in various computer vision tasks [13], [14]. In the context of human detection, DNN-based methods have been shown to outperform traditional methods in terms of both accuracy and speed. There are many successful models in this area, such as YOLO [15], Faster R-CNN [16], and SSD [17]. Among them, YOLO is widely used because of its fast speed and high accuracy. In [18], a lightweight version of YOLO, YOLOv3-tiny, was proposed for vehicle and pedestrian detection. The model was optimized for real-time performance and achieved high accuracy on pedestrian detection tasks. Another interesting approach is using DNNs to fuse multispectral images for human detection. In [19], a deep convolutional neural network was proposed to fuse RGB and thermal images for pedestrian detection. This method used a multiscale feature extraction network and achieved high accuracy on pedestrian detection tasks. The results showed that the method achieved better detection accuracy than that of using each sensor alone. Overall, the fusion of data from multiple sensors can improve the accuracy and robustness of object detection, while DNNs are able to achieve state-of-the-art performance in terms of both accuracy and speed.

Given the aforementioned challenges and reasons, this paper proposes a human detection strategy based on the fusion of IR and thermal images that can be incorporated into detection and rescue systems, such as Drones or Firefighting Robots. As indicated in Fig. 2, our implementation hardware comprises an NVIDIA Jetson Nano computer, an IMX219-77 IR Camera integrated with two infrared lights to enhance the visibility in dark scenes, and a PureThermal Mini Pro JST-SR Camera (with FLIR Lepton 3.5). Our primary contributions are divided into three aspects. In order to ensure that the cameras can be fused successfully, we first calibrated both cameras using the Light Heating Chessboard technique and aligned the images with appropriate key points. Then, to gather data for training and testing, a suppositional scenario of people calling for help while surrounded by dense smoke was developed. Finally, we customized a lightweight deep neural network based on YOLOv4-Tiny [20] and employed IR and Thermal camera fusion method at the neck of the network for efficient human detection. The experimental results showed that the proposed system can achieve real-time response and the mAP@0.5 accuracy of up to 95%.

The remainder of the paper is given as follows. Section II describes the proposed strategy, which includes a lightweight deep neural network designing and camera calibration methods. The experiments including the dataset collection and editing, the training procedure, and the experimental results are presented in Section III. Finally, Section IV concludes the paper and discusses potential research directions in the future.

## II. PROPOSED METHOD

In this section, we discuss the methods adopted to synchronize the data captured by IR and thermal cameras. Following that, we introduce a multi-sensor feature fusion strategy designed based on the YOLOv4-Tiny network architecture, a well-known lightweight deep learning model.

### A. Geometric Camera Calibration

The Light Heating Chessboard technique is a two-in-one geometric camera calibration method that requires exposing the chessboard to sunlight as described in Fig. 3a. We can adjust IR cameras using the same methods as with regular visible cameras. As for the case when capturing pictures with a thermal camera, there is a contrast between the black and white squares due to unequal heat absorption of different colors. This technique is crucial to ensure that the image points $(u_c, v_c)$ are appropriately mapped to their corresponding real-world coordinates $(x_w, y_w, z_w)$ as the formulas below:

$$\begin{bmatrix} u_c \\ v_c \\ 1 \end{bmatrix} = A[R_c | T_c] = \begin{bmatrix} f_x & 0 & c_x \\ 0 & f_y & c_y \\ 0 & 0 & 1 \end{bmatrix}[R_c | T_c] = \begin{bmatrix} x_w \\ y_w \\ z_w \\ 1 \end{bmatrix} \quad (1)$$

After gathering a number of photos of the chessboard with the thermal and IR cameras, we use the Camera Calibration Toolbox of MATLAB to estimate the intrinsic matrix and distortion coefficient. The intrinsic matrix or camera matrix $(A_c)$ contains the information of internal parameters of the camera, such as the focal length and principal point. The distortion coefficient corrects any lens distortion that may be present in the images. There are two basic kinds of distortion with total 5 variables, including radial distortion $(k_1, k_2, k_3)$ and tangential distortion $(p_1, p_2)$. The formula below, where $(x, y)$ is the input picture size, represents both types of distortions:

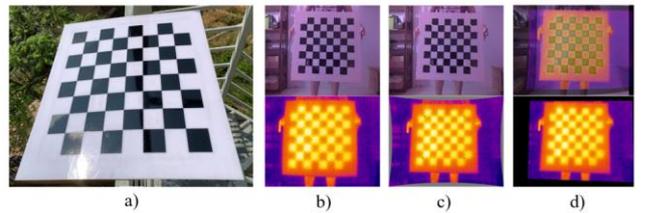

Fig. 3. Image alignment procedure: a) Chessboard is heated by sunlight, b) Chessboard original image, c) Undistortion images, d) Perspective transformed image

$$\begin{cases} r = \sqrt{x^2 + y^2} \\ x_{Radial} = x(1+k_1r^2+k_2r^4+k_3r^6) \\ y_{Radial} = y(1+k_1r^2+k_2r^4+k_3r^6) \\ x_{Tangential} = x + \left[2p_1xy + p_2\left(r^2+2x^2\right)\right] \\ y_{Tangential} = y + \left[p_1\left(r^2+2y^2\right)+2p_2xy\right] \end{cases} \quad (2)$$

*B. Image Alignment*

The aligned image refers to an image that has been processed and transformed to match the position, orientation, and scale of another image. We selected the IR image with the longer focal length as the foundation and the thermal image was transformed so that they overlap together. This approach helps us to unify the dataset since we just need to determine only one set of bounding box coordinates of objects in each scene and assures that the object is present in both the thermal and IR images. Choosing four pairs of corresponding points from every four corners in the chessboard of the undistorted camera image is the first step in the procedure. We employed the following calculation formula to obtain the homography matrix using these corresponding locations:

$$s \begin{bmatrix} x' \\ y' \\ 1 \end{bmatrix} = H \begin{bmatrix} x \\ y \\ 1 \end{bmatrix} = \begin{bmatrix} h_{11} & h_{12} & h_{13} \\ h_{21} & h_{22} & h_{23} \\ h_{31} & h_{32} & h_{33} \end{bmatrix} \begin{bmatrix} x \\ y \\ 1 \end{bmatrix} \quad (3)$$

The homography matrix is a 3x3 transformation matrix that maps points from one image to their corresponding points in the other image. With the aid of this matrix, we can conduct a perspective transformation on the thermal image in order to have it partially aligned with the IR image. All the procedures for the resulting image alignment are given in Fig. 3.

*C. YOLOv4-Tiny-based Feature Fusion Deep Neural Network*

The concept of feature fusion combines features from numerous sources to enhance performance in a variety of computer vision tasks. YOLOv4-Tiny is a variant of the popular YOLO (You Only Look Once) object detection model that uses a compact and more efficient network architecture, making it ideal for real-time applications [20]. We modified the YOLOv4-Tiny network to carry out feature fusion with the inclusion of two distinct kinds of input images, both with dimensions of 416x416. For the objective of extracting features, these images are passed through the CSPdarknet53-Tiny backbone network [21], [20], [22]. However, we employ the Mish function [23] as the activation function instead of the Leaky ReLU function. In order to keep the model from overfitting, the Mish function has the ability to self-adjust the magnitude of the derivative. Other studies also show that the Mish function offers faster convergence and greater accuracy when compared to the Leaky ReLU function due to its smooth curve being more continuous. The feature pyramid network (FPN) structure is utilized at the neck of the architecture to create multi-scale feature pyramids which increase the feature representation of images.

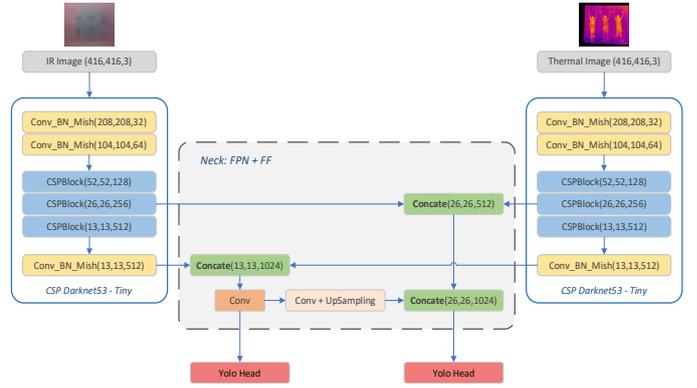

Fig. 4. Deep neural network architecture with IR and thermal feature fusion

Our proposed fusion method also takes place in the part by interfering with a concatenate layer to combine two pairs of informative matrices with dimensions of 26x26 and 13x13. These outcomes are then fed to the YOLO model's head to classify and predict bounding boxes. Even though two images are passed through the input, we only use a single bounding box coordinate because we previously cropped the redundant parts of both images to bring them to the same center and image size. The completely proposed lightweight network architecture is illustrated in Fig .4

In our modified network, the loss function is kept identical to that in the original model. Multiple losses, including localization loss, confidence loss, and class loss, are combined. The localization loss measures the difference between the predicted bounding box and the ground truth bounding box. The confidence loss determines how confident the model is in its prediction of an object being present in the bounding box. The class loss finds out the difference between the predicted class probabilities and the ground truth class probabilities. The final loss is a weighted sum of the above three loss terms.

Adam optimizer [24] is typically used in YOLOv4-tiny. Adam is a popular gradient-based optimization algorithm that brings together the advantages of the AdaGrad and RMSProp algorithms. It uses adaptive learning rates to update the parameters of the model during training, which can help accelerate convergence and improve the overall performance of the model. After integrating all the revisions, our network architecture has a total of approximately 6 million parameters.

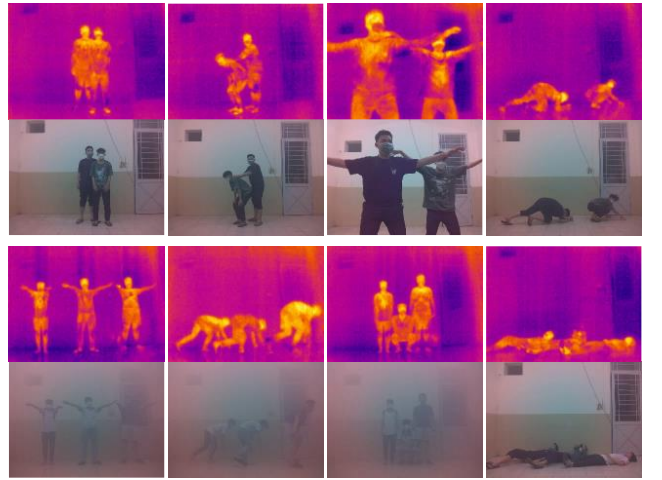

Fig. 5. The diversity of datasets depicting persons in smoke.

## III. EXPERIMENTS AND RESULT

### A. Dataset and Training

In our study, we aimed to create a diverse and generalized dataset for the purpose of simulating fire situations. To achieve this, we staged various scenarios by releasing white smoke into a closed room and positioning people in different ways to symbolize people asking for assistance. Our collection of images contains a variety of situations, such as changing the number of people in the frame, their body poses, and the camera's distances and angles of view. In order to avoid the model from inferring information based solely on the color of clothing, we also made sure that we altered the outfits of the people in the picture.

A total of 3,000 photos make up our dataset, with half of them captured by IR camera and the rest by the thermal camera. To ensure that corresponding images from the two cameras were captured at the same time, we used the Python threading programming technique during the collection process. With the camera matrix and distortion coefficients discovered during camera calibration, photos were undistorted. The thermal image was subsequently transformed using the wrapPerspective function of OpenCV and the homography matrix we computed using the image alignment approach so that when we placed it over the IR image, the two images would overlap. Following that, we went on to label the photos, only labeling the IR images and mapping them to the associated Thermal images because both images are basically identical. Several illustrations of our dataset can be seen in Fig. 5.

The training was carried out by a laptop equipped with a Core-i7 processor running at 2.60 GHz, 8GB of RAM, and an integrated NVIDIA GeForce GTX 1650 with 4GB of RAM. We used hyper-parameters such as 100 epochs, batch size of 32, and learning rate of 0.003. The training procedure took a total of two hours.

### B. Experimental Results

For human detection in real-time, we present the results and comparisons of our proposed method. As demonstrated in Fig. 6, the model can detect humans in a variety of positions and postures, and it achieves excellent detection accuracy even in the presence of occlusions or overlapping individuals.

To evaluate the performance of the proposed method, we use a validation set to compute several evaluation metrics as shown in Table I, including average IoU, precision, recall, F1-Score, and mAP@0.5. To further assess the performance of our method, we run it on a separate test set and the results are summarized in Table II. Using each single camera, we also compare our method with YOLOv4-tiny and YOLOv5s, one of the lightest versions of YOLOv5. The comparison criteria include mAP@0.5, mAP@0.5:0.95, and frame-per-second (FPS). Our findings demonstrate that our method outperforms both YOLOv4-tiny and YOLOv5s in terms of mAP@0.5 and mAP@0.5:0.95 on the test dataset. Specifically, our method achieves a significantly higher level of accuracy, with improvements ranging from 3 to 38 percent compared to the baseline models. These results highlight the superior performance of our approach in accurately detecting and localizing objects in the examined dataset. When considering execution speed, our solution operates slightly slower than YOLOv4-tiny, resulting in a decrease of 3 FPS. However, in comparison to YOLOv5s, our method performs competitively at the same speed of 15 FPS. This indicates that our solution strikes a balance between accuracy and speed, offering an effective trade-off for real-time object detection tasks.

## IV. CONCLUSION

In this paper, we present a human detection system that can be used in search and rescue situations with heavy smoke. By utilizing the proposed feature fusion strategies based on deep learning, human detection accuracy can be enhanced in such difficult circumstances. This crucial study can greatly enhance the chances of people surviving in an emergency fire. This work has the potential to lead to the integration of the technology into unmanned rescue vehicles, such as drones and autonomous cars. Future research can concentrate on enhancing the execution time and increasing the application situations of the proposed technology.

TABLE I. MODEL TRAINING RESULTS

| Avg. IoU | Recall | Precision | F1-Score | mAP@0.50 |
|---|---|---|---|---|
| 71.89 % | 0.94 | 0.89 | 0.91 | 96.97 |

TABLE II. PERFORMANCE COMPARISON

| Models | Image Types | | Metrics | | |
|---|---|---|---|---|---|
| | Thermal | IR | mAP@0.5 | mAP@0.5:0.95 | FPS |
| YOLOv4-Tiny | ✓ | | 87.98 | 68.32 | **18** |
| | | ✓ | 73.32 | 42.85 | **18** |
| YOLOv5s | ✓ | | 91.37 | 77.45 | 15 |
| | | ✓ | 79.45 | 63.84 | 15 |
| **Ours** | ✓ | ✓ | **95.92** | **80.23** | 15 |

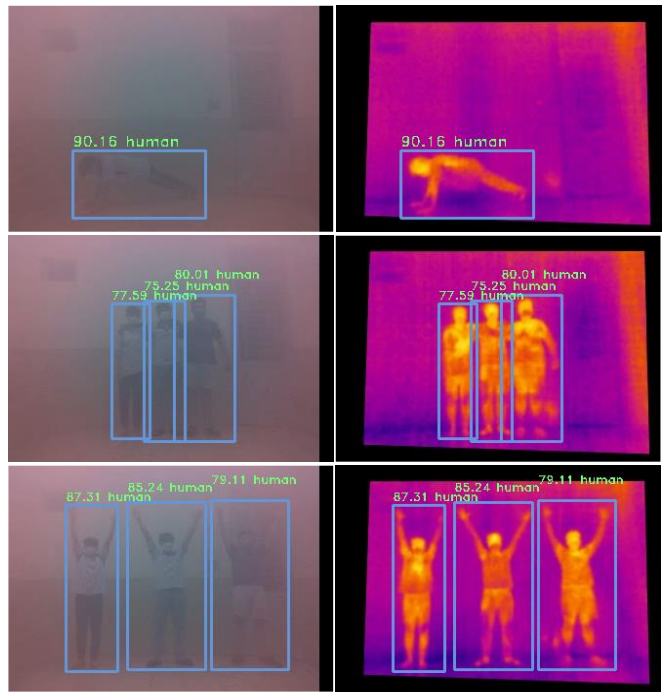

Fig. 6. Human detection implementation in a dense smoke environment with NVIDIA Jetson Nano (better to see in zoom)